\documentclass{article} 

\usepackage{nips10submit_e_mod,times,amsmath}
\usepackage{graphicx}
\usepackage{wrapfig}

\usepackage{amsmath}
\usepackage{algorithm}
\usepackage{algpseudocode}

\title{Dyna Planning using a Feature Based Generative Model}

\author{
Ryan Faulkner\\
School of Computer Science\\
McGill University\\
Montreal, QC, H3A2A7 \\
\texttt{ryan.faulkner@mail.mcgill.ca} \\
\And
Doina Precup \\
School of Computer Science \\
McGill University\\
Montreal, QC, H3A2A7 \\
\texttt{dprecup@cs.mcgill.ca} \\
}

\begin{document}

\algrenewcommand{\algorithmiccomment}[1]{\hskip3em$\%$ #1}

\maketitle

\begin{abstract}
Dyna-style reinforcement learning is a powerful approach for problems where not much real data is available. The main idea is to supplement real trajectories, or sequences of sampled states over time, with simulated ones sampled from a learned model of the environment. However, in large state spaces, the problem of learning a good generative model of the environment has been open so far. We propose to use deep belief networks to learn an environment model for use in Dyna. We present our approach and validate it empirically on problems where the state observations consist of images. Our results demonstrate that using deep belief networks, which are full generative models, significantly outperforms the use of linear expectation models, proposed in Sutton et al. (2008).
\end{abstract}

\section{Introduction}

Recent reinforcement learning (RL) research is devoted to increasingly large problems, in which an agent is forced to learn quickly, with a small amount of data.  One approach that attempts to make efficient use of the existing data 
is the Dyna architecture proposed by Sutton (1990)\nocite{sutton1990first}.  In this case, data obtained by an agent from its interaction with the environment is used to learn both a value function (which guides the action selection) and a generative model of the environment.  The model can then be used to ``fantasize'' additional data, which the agent can use to improve its value estimates.  If the model is of good quality, the sampled data will be very useful as a supplement for real data.  This approach has been quite successful as a planning and learning tool in discrete domains, where a multinomial distribution can easily be used for the model.  However, in continuous or very large state spaces, which require function approximation, the issue of building a good model has remained elusive.

In this paper, we propose to use deep belief networks (Hinton et al., 2006)\nocite{hinton2006reducing} to learn an environment model for a Dyna architecture.
Our goal is to take a step towards using such architectures to tackle reinforcement learning in environments with very large state or observation spaces, consisting for example of images, sounds or rich sensor readings.  Deep belief networks have proven to be very effective at learning how to represent such complex data.  However, only a few applications have been implemented using deep belief nets for the analysis of temporal data (e.g. Taylor et al., 2007\nocite{taylor2007modeling}, Memisevic \& Hinton, 2009\nocite{memisevic2010learning}).  Our proposed approach is to learn a  deep representation for the states of the environment first.  Then, a temporal layer connecting two deep belief nets is learned; this will model the conditional distribution of the next state, given the current state (See \S\ref{sec:algorithm}).  In the current implementation, models are learned separately for each action as the space of actions is discrete and small in number in our experiments.  Once the action models are learned, they can be used to generate samples of states, by setting the input as the sensor reading of the current state, and performing inference to generate a next state.  These samples can be used both for extra learning steps, as well as in order to plan which action should be chosen.  We mainly investigate the learning aspect in this paper.  We note that free-energy models have been used before in reinforcement learning to represent value functions (Sallans \& Hinton, 2004\nocite{sallans2004factored}) or policies (Otsuka, 2010\nocite{otsuka2010thesis}).  However, to our knowledge, this is the first time they are used as a generative model in the Dyna framework - a setting where the full power of such a model can be exploited best.


\section{Background}\label{sec:background}



In reinforcement learning, an agent interacts with its environment at discrete time steps $t=0, 1, \dots$.  At each time step, the agent receives information about the current state $s_t$, chooses an action $a_t$ and the environment transitions to a new state $s_{t+1}$, emitting a reward $r_{t+1}$.  In general, learning takes place on-line; the agent attempts to estimate a policy $\pi$, mapping states to actions, which achieves high cumulative reward.  As an intermediate step, many reinforcement learning methods will estimate a {\em value function}, i.e. an expectation of the cumulative reward.  Value functions can be associated either with states or with state-action pairs.  The state value function for a fixed policy $\pi$ is defined as:
\[
V^{\pi}(s) = E_{\pi}[r_{t+1} + \gamma r_{t+2} + \dots|s_t=s]
\]
where $\gamma\in(0,1)$ is a discount factor, used to de-emphasize rewards that will be received in the distant future.
If the environment is Markovian, it is guaranteed to have a unique optimal state value function:
\[
V^*(s) = \max_{\pi} V^{\pi}(s)
\]
which can be achieved by at least one, deterministic policy.  State-action value functions can be defined similarly (see Sutton \& Barto, 1998, for a comprehensive description).

If the agent follows a fixed policy,  temporal difference (TD) learning (Sutton, 1988)\nocite{sutton1988td} provides a way to use experience to estimate the value function. For example, in TD(0), the following update is performed at every time step $t$:
\begin{equation*}
V(s_t) \leftarrow V(s_t) + \alpha[r_{t+1} + \gamma V(s_{t+1}) - V(s_t)] 
\end{equation*}
where $\alpha\in (0,1)$ is a learning rate parameter.
In domains with many states, $V$ can be represented by a function approximator with parameter vector $\vec{\theta}$.  In this case, the TD(0) update rule becomes:
\begin{equation*}
	\vec{\theta}_{t+1} = \vec{\theta}_t + \alpha[r_{t+1} + \gamma V_t(s_{t+1}) - V_t(s_t)] \nabla_{\vec{\theta}_t}V_t(s_t) \;
\end{equation*}
where $\vec{\theta}_t$ denotes the estimate of the parameter vector at $t$, and $V_t$ is a shorthand for $V_{\vec{\theta}_t}$.
The expression of the gradient, $\nabla_{\vec{\theta_t}}V_t(s_t)$, depends on the type of function approximator used (e.g. linear, neural network etc.)

If the goal is to learn an optimal control policy, the Q-learning algorithm can be used to estimate the state-action value function, with update rules similar to the ones above (see Sutton \& Barto, 1998, for details).  When function approximation must be used, a separate approximator is usually maintained for each action when the space of actions is discrete and when the number of actions is not so large that this approach is prohibitive.  These conditions are met for the the experiments that follow.    

\begin{wrapfigure}{r}{0.5\textwidth}
\centering
\includegraphics[height=3.0cm]{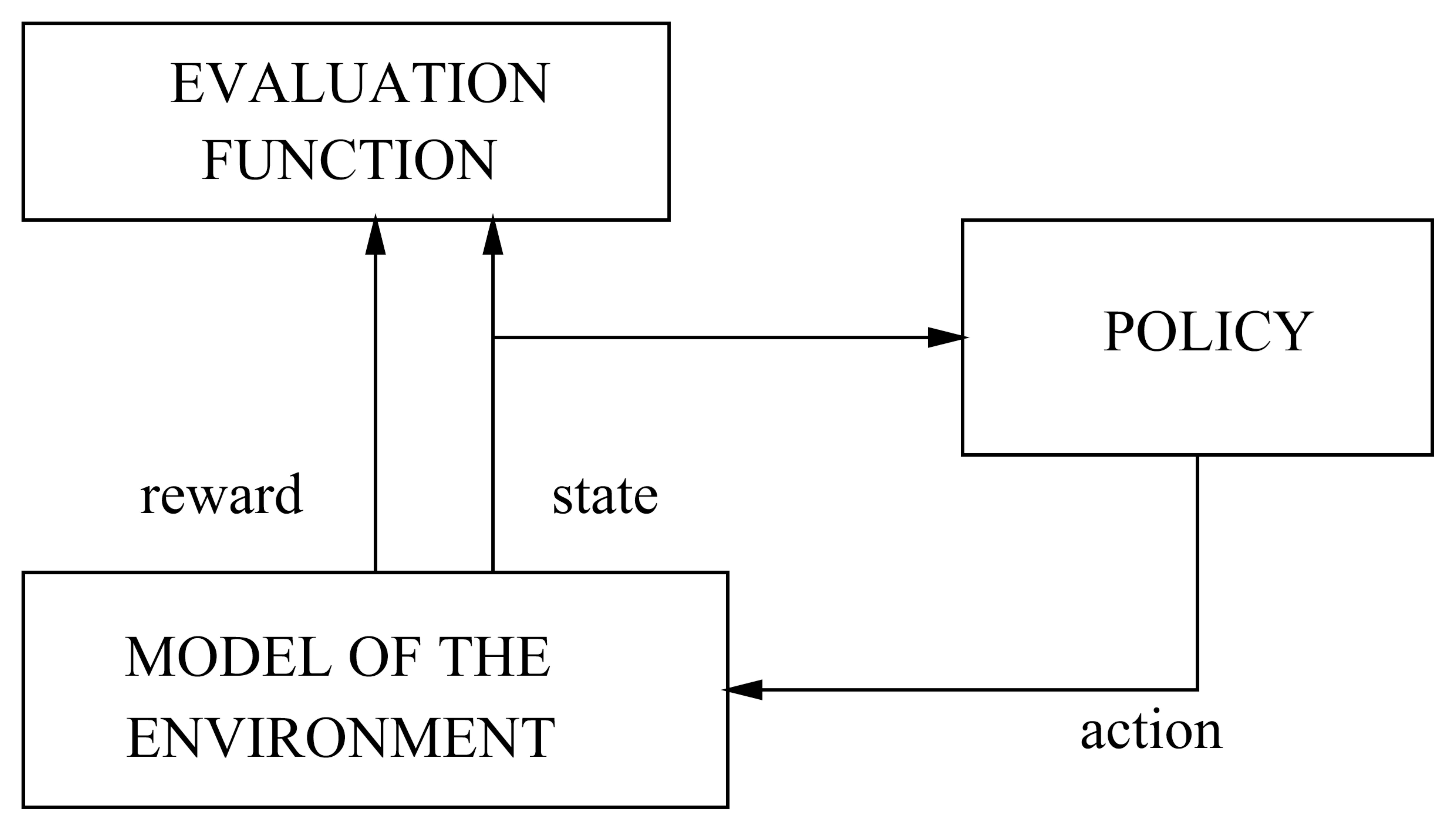}
\caption{Dyna Architecture}
\label{fig:dynafig}
\end{wrapfigure}

In some applications, for example robotics or active vision, it is difficult to obtain a sufficient amount of real data by interacting with the environment directly, because generating experience may be risky or time consuming.  
The Dyna architecture is a framework for planning and learning introduced by Sutton (1990) and intended for this situation.    The architecture is presented in Figure \ref{fig:dynafig}.  The main idea is that the agent will use real experience to build a ``simulation" (i.e. generative) model of the environment, which can then be used to sample new transitions.  For every step of real experience, the agent will also sample transitions and rewards from its learned model; these are used just like the real experience, to perform TD or Q-learning updates.  Dyna has proven successful in large discrete problems, where experience is scarce; however, for large problems, where function approximation must be used, finding a good way to model the environment has been difficult.  Recently, Sutton et al. (2008) proposed an approach that can be used for linear function approximation, in which only ``expectation'' models are used, instead of fully generative models.  In this case, the model consists of learning the expected value of the next feature vector, $\phi(s_{t+1})$, given the current feature vector $\phi(s_t)$.  They show that planning steps can be performed correctly with such a model, and that the model can be used in combination with linear function approximation.  However, this approach does not work with neural networks or other non-linear function approximators.

\section{Proposed approach}
\label{sec:algorithm}

Our goal in this paper is to leverage the representational power of deep belief networks in order to be able to learn generative models of environments used in reinforcement learning tasks.  
We use the approach taken by Hinton et al. (2006)\nocite{hinton2006fast} to greedily train a deep network layer-by-layer by training a Restricted Boltzmann Machine (RBM) (Smolensky, 1986)\nocite{smolensky1986information} at each layer.
Such models can then be used in a Dyna-style architecture, in order to learn good value functions and policies; we note that, while we explore mainly Dyna in this paper, other uses of the models are also possible (e.g., in order to estimate the risk of certain policies, based on the variance of the samples obtained using the model).

\begin{wrapfigure}{r}{0.5\textwidth}
\vspace{-20pt}
\centering
\includegraphics[height=5.0cm]{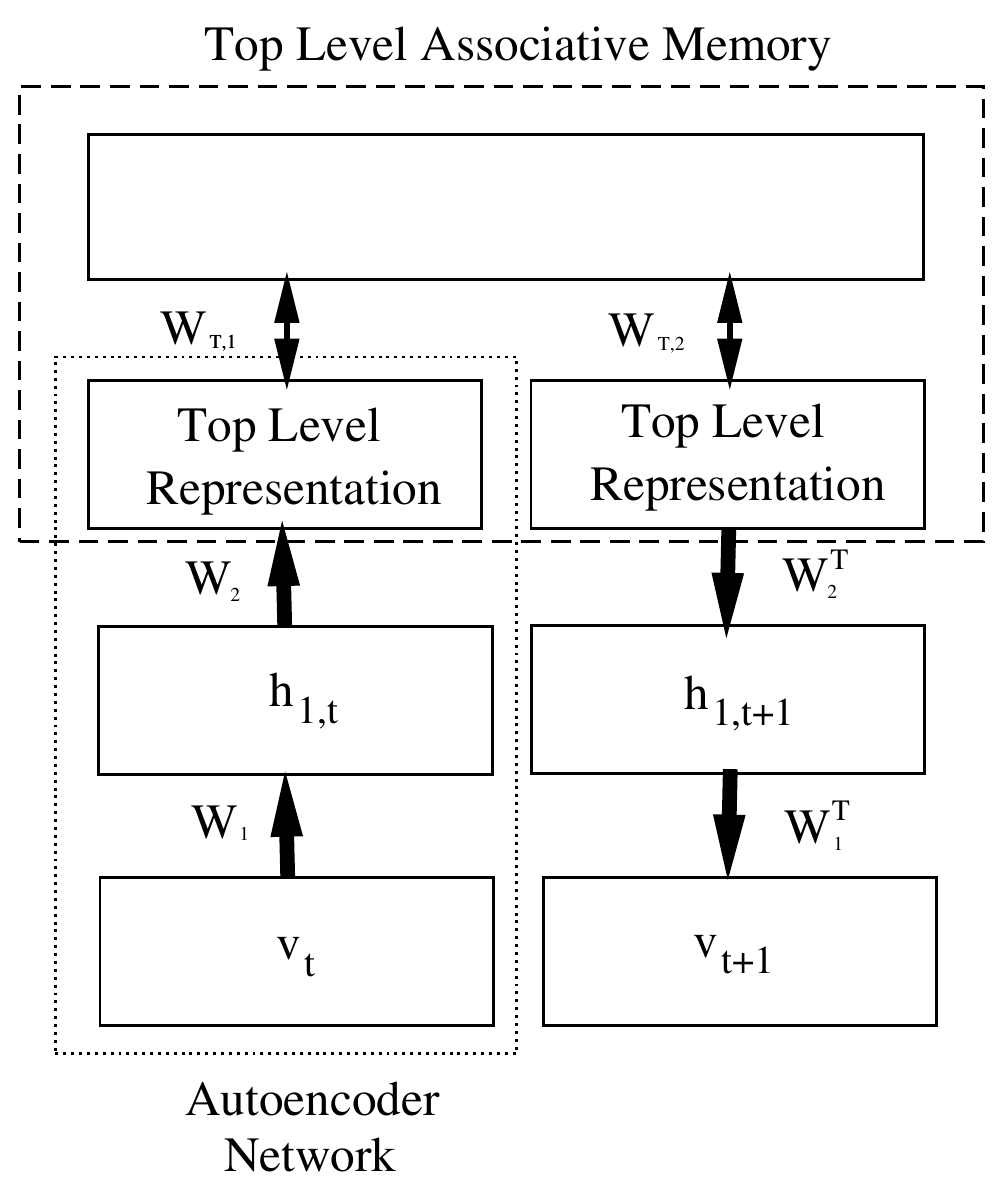}
\caption{The generative network used for Dyna learning. \textbf{W} represents the weights at each layer of the deep network.  The visible units are represented by the bottom layer of variables in the autoencoder network with hidden layer representations up to the top-level of the auto-encoder.}
\label{fig:fullModel}
\vspace{-40pt}
\end{wrapfigure}

Figure \ref{fig:fullModel} illustrates the deep belief model we use.  It is composed of two parts: an autoencoder network (duplicated on the left and right side) and an associative memory (RBM) at the top level.  The autoencoder network is used to generate a good, compact representation of the state, which summarizes well the observed data; this is similar to the use of deep networks for unsupervised learning, summarized above.  The RBM at the top is the \emph{temporal layer} of the model; its role is to model the temporal dependencies between states (assuming that, at the level of the hidden variables, the environment is well represented as a Markov Process).   The topology closely resembles the deep network discussed by Hinton et al. (2006)\nocite{hinton2006fast}.  It differs from other forms of deep networks based on energy models that are used for prediction over time (Taylor et al., 2007; Sutskever et al., 2008)\nocite{taylor2007modeling,sutskever2008recurrent} because it focuses first on reducing the dimensionality of the data, and only uses this reduced latent variable representation in the temporal dependency model.  Using a lower dimensional representation makes inference and learning the temporal dynamics of the environment much faster, due to the potential representational power over the higher level units (Hinton et al., 2006; Bengio, 2009)\nocite{hinton2006fast}.  This is important in our case, because fast inference is essential in problems in which an agent must take actions in a timely manner.  

The learning algorithm has two phases: \emph{training the model} and \emph{learning the value function and policies}, using states generated by the model to supplement real experience.  We now describe each of these components.

\subsection{Training, Prediction \& Dyna}\label{sec:genPrediction}

Training the network involves learning each layer independently as described in the previous section.  The data consists of a time sequence of environment states, $s_1, s_2,\dots$.  A separate model is learned for each action.  If the time series is generated according to an exploration policy, the samples $s_t, a, s_{t+1}$ corresponding to a particular action $a$ will be used to train the model for $a$.
The autoencoders are learned first, using greedy layer-wise training (Hinton et al. 2006)\nocite{hinton2006fast}.  Note that this step does not require a time sequence, as the autoencoder is aimed at representing states in a more compact way.

Once the autoencoders are learned, the
 parameters of the temporal layer must be trained.  Each pair of states $s_t$ and $s_{t+1}$ is used as data at the visible units.  This evidence is propagated up in the autoencoders, to obtain corresponding latent variable representations $h_t$ and $h_{t+1}$.  These are used as evidence to train the 
 temporal layer of the model, using the standard RBM procedure described by Hinton et al. (2006)\nocite{hinton2006fast}.


When the full model has been trained, it can be used to generate samples of future states, given a current state observation.  
The current state $s_t$ represents the data.  Its high-level representation, $h_t$, is generated as described above. 
  Gibbs sampling is performed over the temporal layer, holding the visible units of the RBM corresponding to the representation $h_t$ fixed, and allowing the remaining temporal layer units to settle to their equilibrium distribution.  This effectively yields an unbiased sample from $P(h_{t+1} | h_t)$.  Once the Gibbs sampling is finished and $h_{t+1}$ has been determined, 
 the low-level representation (corresponding to the observation of the next state, $s_{t+1}$) is determined by performing a generative downward pass over the autoencoder model.

Using these predictions it is possible to generate $K$-step trajectories (Sutton et al. 2008)\nocite{sutton2008dyna} of simulated experience for every real transition.  These simulated trajectories are rooted at observed states in order to ensure that the relative probabilities with which states are observed in the data are preserved, to the extent possible.  Note that these probabilities are crucial in reinforcement learning, and they vary as the agent's policy changes.  Ideally, the models should be trained continually, as the agent's policy changes, or importance sampling should be incorporated in the inference procedure to make sure that the state visitation distribution is tracked correctly.  However, for simplicity, we separate the model training phase and the value function training, without any loss in the quality of the results (as we will see in the next section).

Note that the number of simulated transitions $k$ is usually a very important parameter in Dyna-style architectures.  If the model is imprecise, the trajectories drift away from the real distribution as $k$ becomes large, and performance degrades.  Having a good-quality model is crucial for this approach to work.

\section{Experiments \& Results}
\label{sec:experiments}

In order to evaluate our approach, we use navigation tasks, but with a twist: instead of observing the exact identity of the state, the agent observes a high-dimensional vector representing the state.  Observations may be noisy, or they may be uniquely associated with the state.  However, even if the environment is fully observable, function approximation is needed for the agent to correctly estimate returns, because of the dimensionality of the observation vector.

We chose to associate states with images, because this facilitates the visual interpretation of results, and also because image processing with deep belief networks has been explored extensively.  The first domain we used is a small Markov chain built on top of the  MNIST data set \cite{lecun1998gradient}.  The second set of experiments uses larger problems, built on top of the NORB \cite{lecun2004learning} data set, and the goal is to learn optimal control (rather than just estimating the value of a fixed policy).  
The key feature of both domains is that states have a high-dimensional low-level representation, so they may be represented much better using high-level features.  We now present the details of the two domains and discuss our results.

\subsection{MNIST Environment}




\begin{wrapfigure}{r}{0.5\textwidth}
  \centering
  \includegraphics[height=1.5cm]{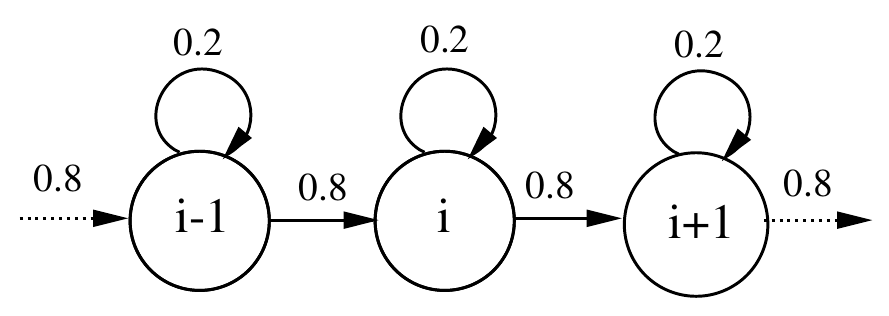} 
  \caption{The general transition diagram of the Markov chain defining the MNIST domain.}
  \label{fig:transitionKernels}
\end{wrapfigure}

The MNIST database, created by Yann Lecun and Corinna Cortes (1998), consists of handwritten decimal digits.  The full data set consists of $60000$ training samples and $10000$ test samples; each sample consists of $28 \times 28$ ($784$) monochrome pixels.  
We defined 10 states, corresponding to the MNIST classes.  The purpose of the MNIST experiment is only to evaluate the effectiveness of our model in the context of Dyna, thus actions are omitted for this toy domain.
We use a simple Markov chain over these states, depicted in Fig.~\ref{fig:transitionKernels}.  The transition probabilities are  $P(s_{t+1} = i + 1 | s_t = i) = 0.8$ and $P(s_{t+1} = i | s_t = i) = 0.2$ for $0 \leq i < 9$.  When $i = 9$, then $P(s_{t+1} = 0 | s_t = 9) = 0.8$ and the probability of staying in state ``9'' is $0.2$.  Only the state corresponding to the class label ``9'' has a reward of $+1$; all other rewards are $0$.  The discount factor is $\gamma=0.9$.  The observation corresponding to a state is sampled uniformly randomly from the corresponding set of digits.  Note that this is actually a partially observable environment, although the observations for a particular state are very closely related.  A similar domain, but on a smaller scale, was used in Otsuka (2010)\nocite{otsuka2010thesis}.

\subsection{Model Learning for the MNIST Domain \& Value Function Learning}

We generated $1000$ transitions from the environment.  To train the autoencoder portion of the network, four levels were used, with $500$, $500$, $200$, and $100$ hidden units respectively (from bottom to top) and $CD_{10}$ - contrastive divergence learning with $10$ Gibbs steps (Carreira-Perpinan \& Hinton, 2005)\nocite{carreira2005contrastive} - trained over $2000$, $1000$, $3000$, and $5000$ sweeps of the data respectively (see Hinton et al., 2006 for details).  The data was also broken into $20$ evenly sized mini-batches.  After training the deep models greedily they were fine tuned using backpropagation with a learning rate of $0.01$ over $20000$ sweeps.  The temporal layer  contained $1000$ hidden units and was trained over $3000$ sweeps with $CD_{20}$ and the data broken into $10$ sequential and evenly sized mini-batches.  Over all training we used a learning rate of $0.005$ with a momentum of $0.9$.  Reconstruction error over the data using just the autoencoder ranged from $4-6$\% on test data after fine tuning.   Training times for the auto-encoder, fine-tuning, and temporal layer ran $4$, $5$, $0.5$ hours respectively. All training was performed on an $8$-core CPU.  


We use this domain as a proof-of-concept for our architecture.  For this purpose, we wanted to evaluate separately errors introduced by the model and errors introduced by the value function estimation.  Hence, we use two ways to represent the value function. 
The first method (\textbf{TAB}) assumes that we have access to the state labels and therefore a table of values can be learned, with a separate entry for each state.   In order to obtain class labels from the representation provided by the deep network, we 
train a three layer neural network (with $500$ and $150$ hidden units), in a supervised manner, using back propagation \cite{lecun1998gradient}.  The classifier was trained on 45000 MNIST training samples for $2500$ sweeps with learning rate $0.1$ and momentum set at $0.5$.  Over a test set of $850$ cases, there were on average $25$ errors, a rate of $2.94$\%, in close agreement with current literature \cite{lecun1998gradient}.  The labels obtained from the classifier are then used as indices in the value function table.

The second method (\textbf{FA}) involves learning a linear function approximator over the features of the state representation (in this case, the MNIST pixel values).  Under this scheme, the value of a state is estimated by averaging the value returned by the function approximator over all of the training data samples for each labeled state.

We used learning rates of $\alpha$ = $0.01$ and $10^6$ updates for \textbf{FA} and $\alpha$ = $0.02$ with $4 \times 10^5$ steps with \textbf{TAB}.  The number of updates to perform was determined empirically by observing when value estimates and the function approximation parameters no longer changed significantly.  We used the Dyna algorithm with $K=5$ (Sutton et al. 2008)\nocite{sutton2008dyna} where sampling of future states was performed with $5000$ Gibbs steps ($2$ hours to generate all samples).   If the final sample of the training data is encountered before the values have converged (within a specified threshold) then another pass over the data is performed.

\subsection{MNIST Results}

\begin{figure}[t]
\vspace{-80pt}
\centerline{
\includegraphics[height=8.5cm]{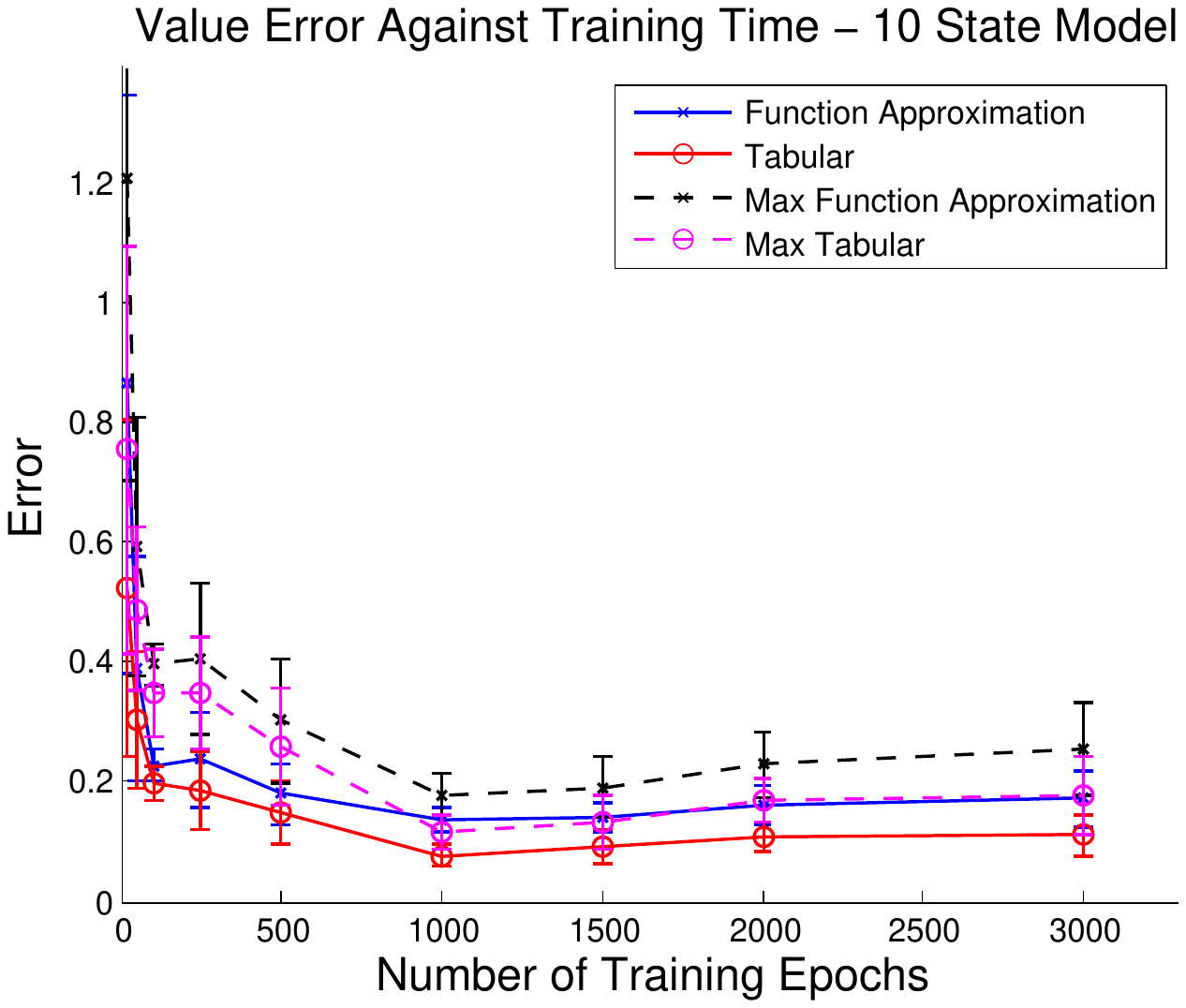} \hspace{-60pt}
\includegraphics[height=8.5cm]{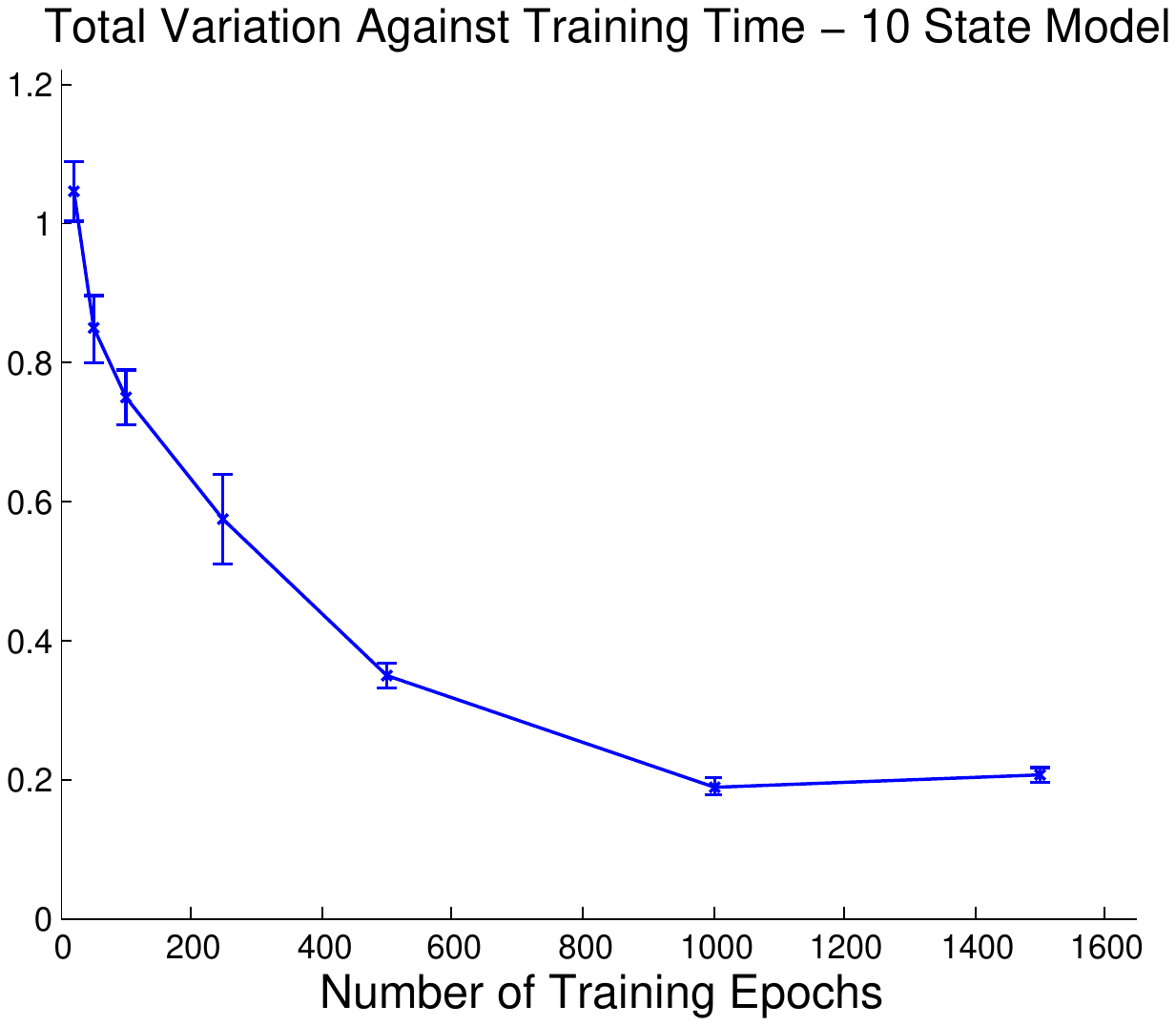} \hspace{-60pt}
\includegraphics[height=8.5cm]{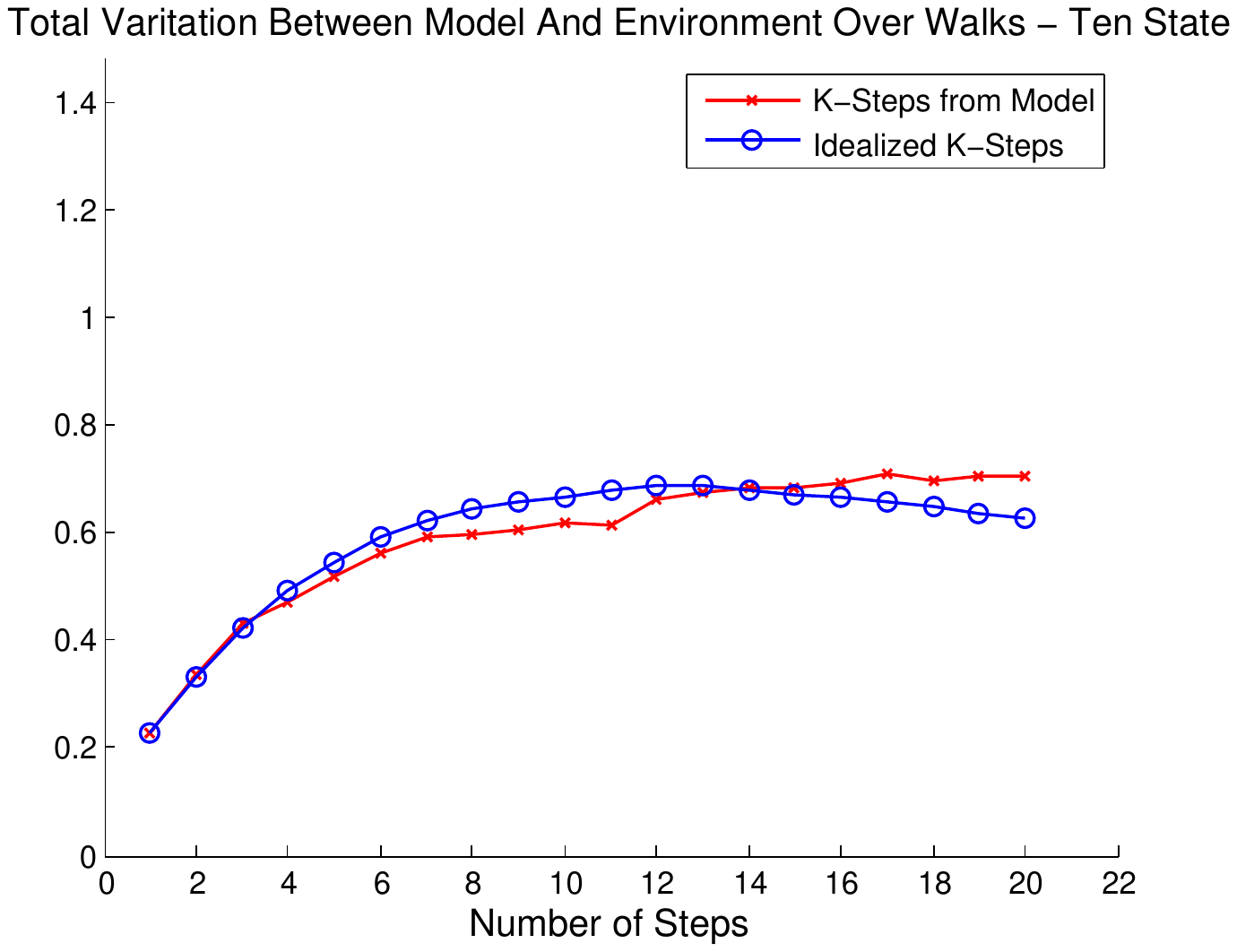} 
}
\vspace{-80pt}
\caption{\small{The error in the value function for the \textbf{TAB} and \textbf{FA} methods and the total variation of the transition matrix sampled from the model and the transition matrix of the environment. At right is the total variation of the simulator distribution over several steps.  }}

\label{fig:MNIST_analytics1}
\vspace{-12pt}
\end{figure}

The values computed under each approach are compared to the true values of the states, which can be computed exactly from the model.  Fig. \ref{fig:MNIST_analytics1} shows the value function error as training progresses (left) as well as the total variation of the deep model compared to the real one (middle panel).
The value error was computed by averaging the absolute differences of value predictions and the true values over the number of states.  The total variation estimates are computed from the transition kernel sampled from model predictions and the true model.  As expected, more training leads to better estimates, both for the model and for the value function.

An important aspect we wanted to test is the ability of the model to project farther into the future.  Therefore, we generated 20-step trajectories over the model, and estimated the total variation between the true $k$-step model of the environment, and the $k$-step model as estimated by the simulation.  The result is presented in Fig.~\ref{fig:MNIST_analytics1} (right).  The plot shows a remarkable ability of the simulated model to track the true model, even on longer trajectories.  These results are very encouraging, as they demonstrate that the models are quite stable.  With this, we now turn to a bigger problem.

\subsection{NORB Environment}



The NORB dataset \cite{lecun2004learning} consists of images of toy models from five object classes: humans, cars, planes, trucks, and four-legged animals.  The toy models are all uniformly coloured white, which means that information about shape is the most relevant indicator of an object's class.

For each class, there are nine distinct models, photographed at a $96 \times 96$ resolution on a blank background from nine different elevations, $18$ azimuths, and under six different levels of illumination.  As the first intended use of NORB was for 3D object recognition, the dataset is composed of stereo pairs over each of the sample configurations, leading to a grand total of $24300$ image pairs.  There are more complex NORB datasets (the set described is referred to as the ``small'' set) which include jittered and cluttered backgrounds, translation of the object from the center, and also distractor objects in the periphery of the image.  Using these image sets is left for future work.

We use a modified version of the ``small'' datset.  Only one of the stereo pairs is used and over the five classes, we chose a single instance, illumination level, and elevation.  This leads to five distinct objects of different classes at 18 azimuths for a total of $90$ samples or states to be used in our RL domain.  To speed up the training of the model we reduced the resolution of the images in a fashion similar to Hinton et al. in \cite{salakhutdinov2009deep}.  The images were first reduced in size to $64 \times 64$ pixels by removing the borders of the samples which contained only background.  The resulting images were then downsampled to a $32 \times 32$ resolution.  

\begin{wrapfigure}{r}{0.5\textwidth}
	
	\vspace{-100pt}
	\centering
	\includegraphics[height=8.5cm]{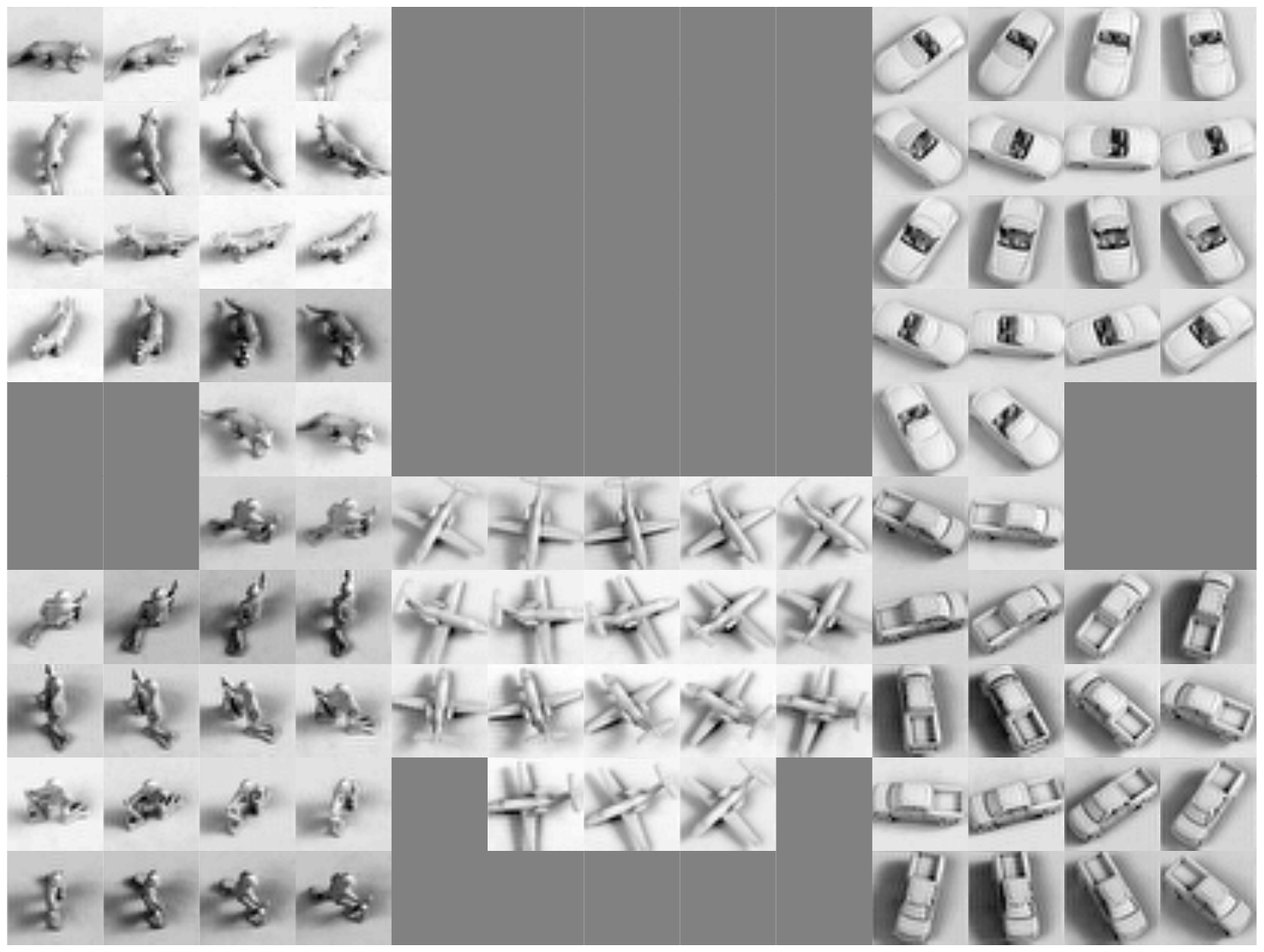} 	
	\vspace{-70pt}
	
	\caption{\small{The NORB environment. The agent moves by taking actions: Up, Down, Left, or Right.}}
  \label{fig:norbMap}
\end{wrapfigure}

Each of the $90$ samples is associated with a distinct state of the environment.  These states are arranged in a 2D map, which the agent is intended to navigate.  The agent has four actions available: up, down, left, and right.  For each action, the agent has a $90\%$ chance of moving in the direction indicated by the action, and a $10\%$ chance of staying in the same state.  If the agent takes an action that would move it off of the map then it remains in the same state deterministically.  The NORB map is shown in Figure \ref{fig:norbMap}.  We intended this environment to simulate a camera moving through an environment in which objects are placed, and the camera views them at different angles, depending on its position relative to the object.

There is a  reward value of $+1$ associated with the top row of car images in the NORB map (top right row) and a reward of $0$ for all other states.  Episodes consist of the agent starting in the top left corner of the NORB map and terminate upon the agent reaching any of the non-zero-reward states.  In order to simulate rewards, a logistic regression model over the predefined state rewards is learned with a learning rate of $0.0001$, over $10^5$ training iterations.  The NORB classes are used as input, and their associated rewards serve as the output.

\subsubsection{The Model for the NORB Environment}

Training the deep model was done using the 90 states as data which, at $32 \times 32$ pixels, required 1024 visible units at the bottom of the network.  Since the NORB data consists of real valued pixel intensities, the first level RBM was trained using a gaussian distribution over the visible units with binary hidden units, similar to the approach taken in \cite{nair20093d}.  For the Gaussian-binary RBM we used $4000$ binary hidden units, a learning rate of $0.00001$, over $20000$ epochs, with $CD_5$, and nine mini-batches ($10$ samples per batch).  The remainder of the deep model was trained with binary units as in the MNIST case.  The binary deep model consists of four levels, with $2000$, $1000$, $200$, and $100$ hidden units, learning rates of $0.001$, $0.001$, $0.001$, and $0.004$ respectively, and training sweeps of $40000$, $20000$, $20000$, and $200000$ respectively.  One batch and $CD_5$ was used over each level.  The learning over the deep model was deterministic, meaning that unit values were not sampled during learning.  Instead, the probabilities were used as the activities over each layer to generate a single representation for each state.  This was acceptable since only one sample is used for each state.  Training the entire auto-encoder portion of the NORB model took roughly $12$ hours for the binary-guassian RBM and one day for the binary deep model.

The data for the temporal layer was generated by choosing actions uniformly randomly, for a total of $7200$ steps.  This data was then partitioned into pairs generated from each action.  The result was roughly $20$ sampled transitions for each state and action, sufficient to model the transitions of the environment.   Note that training samples in this case consist of a pair of top-level representations of NORB states.  In the case of our model, a training sample for the temporal layer consists of $200$ visible binary values, where the first $100$ values represent the state at some time, $s_t$, and the next $100$ values represent the state at the next time step, $s_{t+1}$, sampled from the true MDP defined by the NORB map.  This data was used to train a model for each action where the temporal layer was trained stochastically with 500 binary hidden units, a learning rate of $0.005$, over $2000$ epochs, using $CD_5$, and two batches each containing $900$ samples.  Training time here was about $30$ minutes.  

\begin{wrapfigure}{r}{0.75\textwidth}	
	\vspace{-100pt}
	
  \centerline{
	\includegraphics[height=8.5cm]{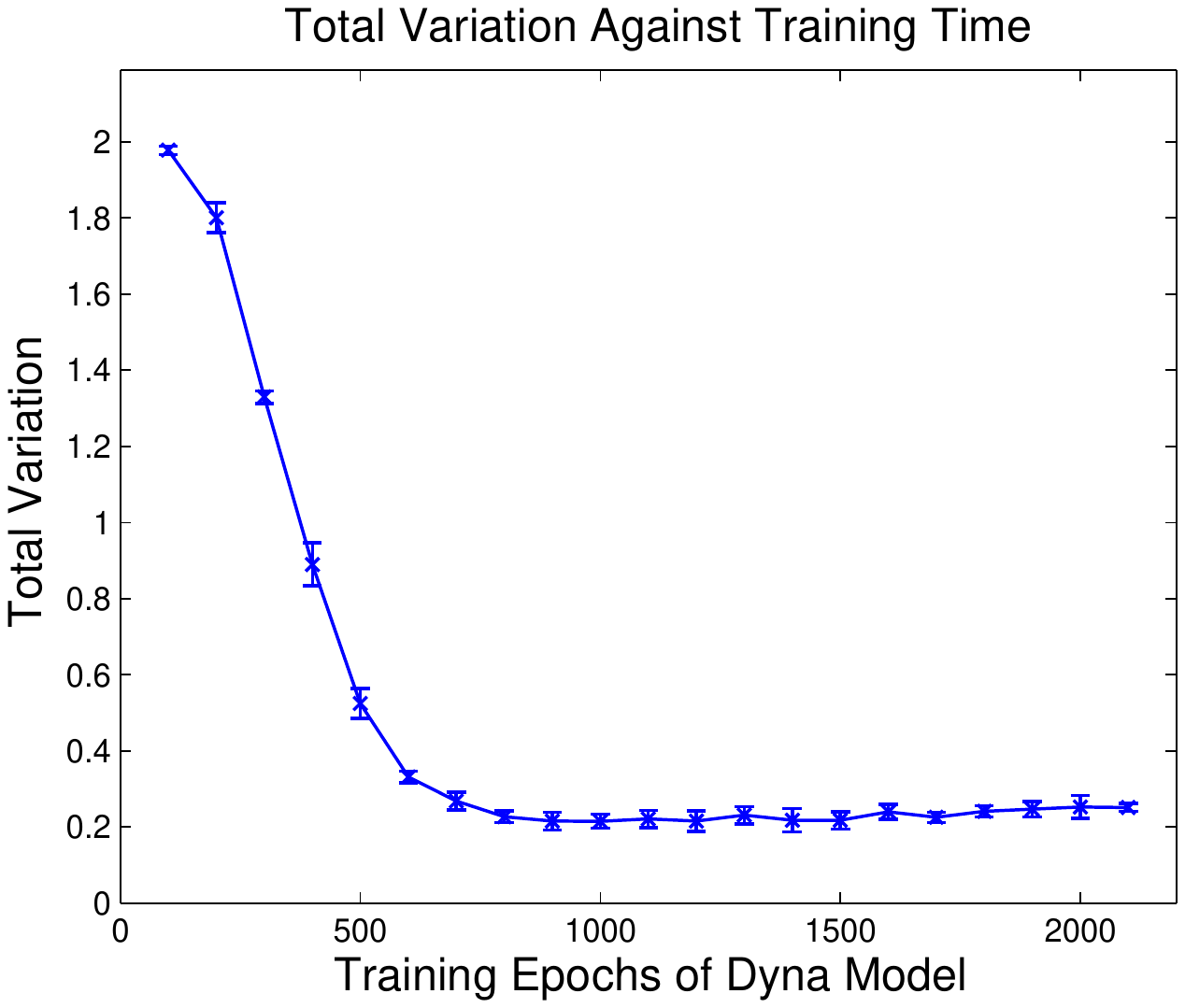} \hspace{-58pt}
	\includegraphics[height=8.5cm]{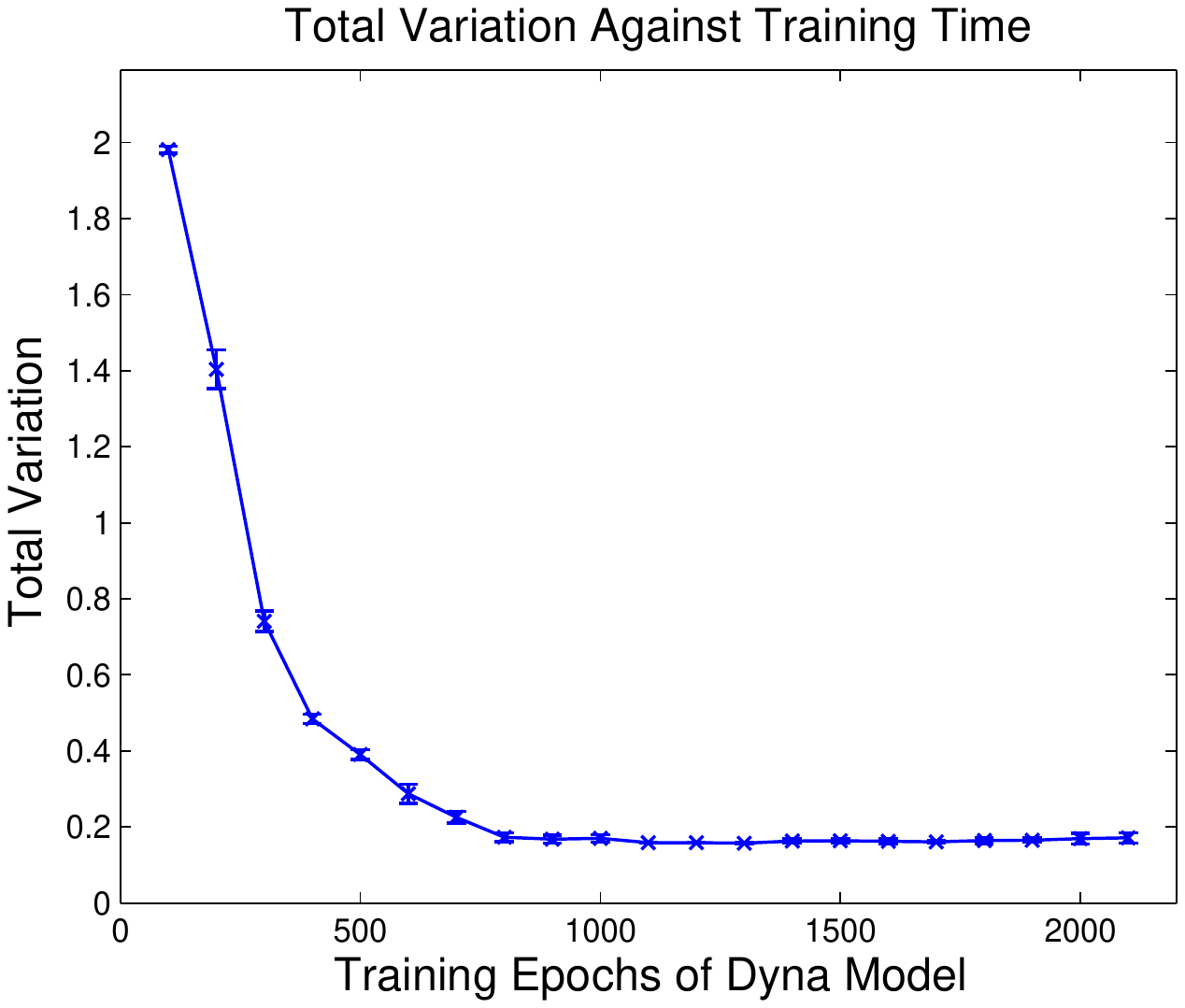} 
	}
		
	\vspace{-80pt}
	
	\caption{\small{The total variation of the NORB models against the true transition function against training time.  The plots show the results for the ``UP'' and ``LEFT'' actions respectively - the remaining actions have similar curves.}  } 
  \label{fig:norbTV} 
\end{wrapfigure}

While training the temporal layer the total variation of the temporal model is sampled after every $150$ epochs of learning.  
To determine the transition function of the model $20$ next step predictions are sampled for each NORB state. The Euclidean distance between each model sample and all of the NORB classes is computed and the class yielding the smallest distance is chosen as the class of the sample.  The total variation between this transition function estimate and the true transition function of the environment is then computed.  Despite the error introduced by the mapping scheme described above clear improvement is observed in the transition function of the model in Figure \ref{fig:norbTV}.

\subsection{Learning Control for the NORB Domain}

To begin learning the value function over the NORB states the agent is initialized in the top left corner of the map then allowed to choose among the available actions.  The environment is episodic and therefore the agent returns to the initial state after visiting a state with reward.  The experiments proceed with a learning rate of $\alpha = 0.001$ and a discount factor $\gamma = 0.9$.  The learning rate over the simulated samples decays to $95\%$ of its value after each episode, but falls no lower than $0.0001$.  Epsilon-greedy exploration is used with an initial value of $\epsilon = 0.9$ that decays to $90\%$ of its value after each episode until it sinks to $0.05$ at which value it remains for all future epsiodes.

\begin{figure}

  \vspace{-80pt}
  \centerline{
	\includegraphics[height=8.0cm]{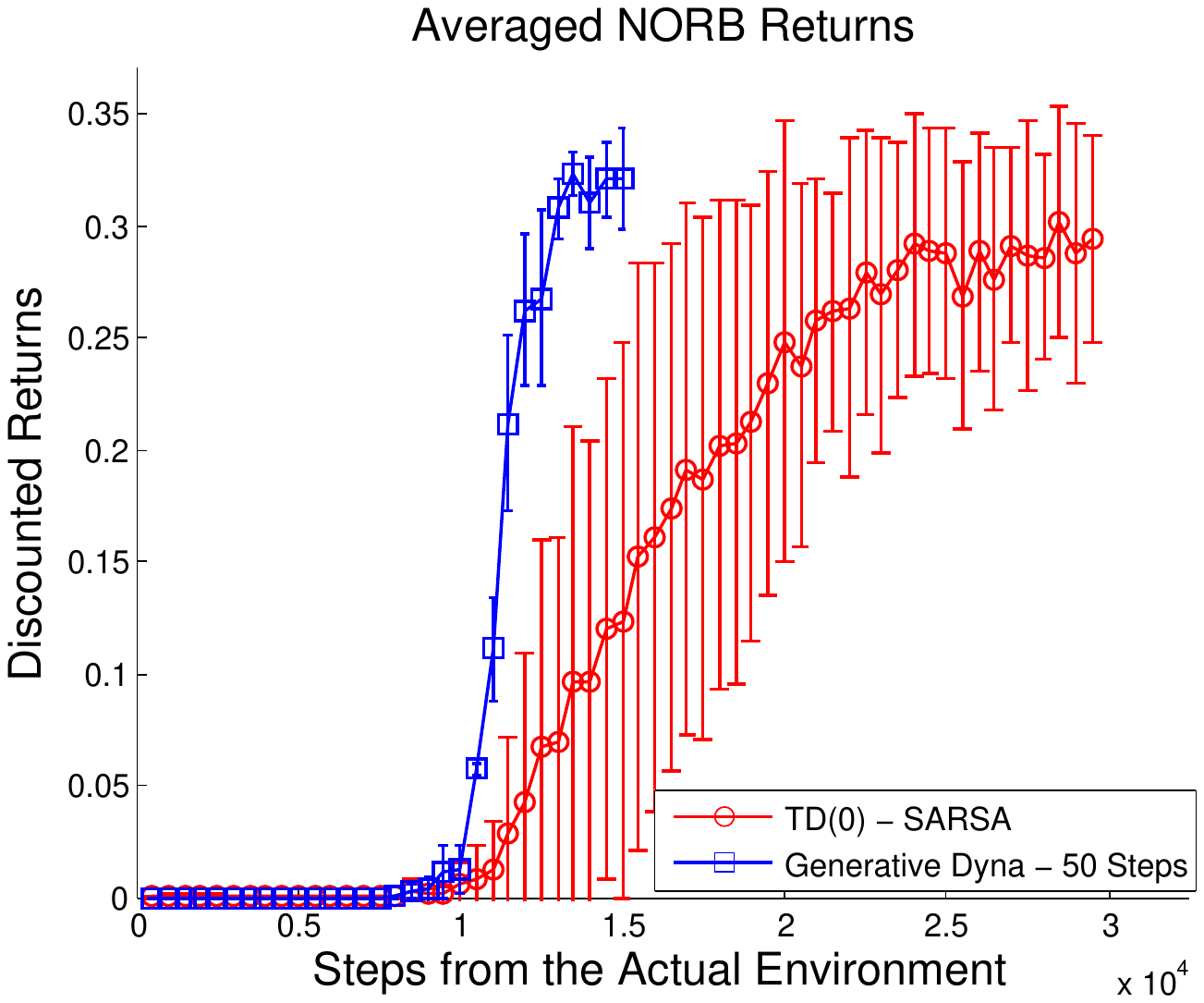} \hspace{-40pt}
	\includegraphics[height=8.0cm]{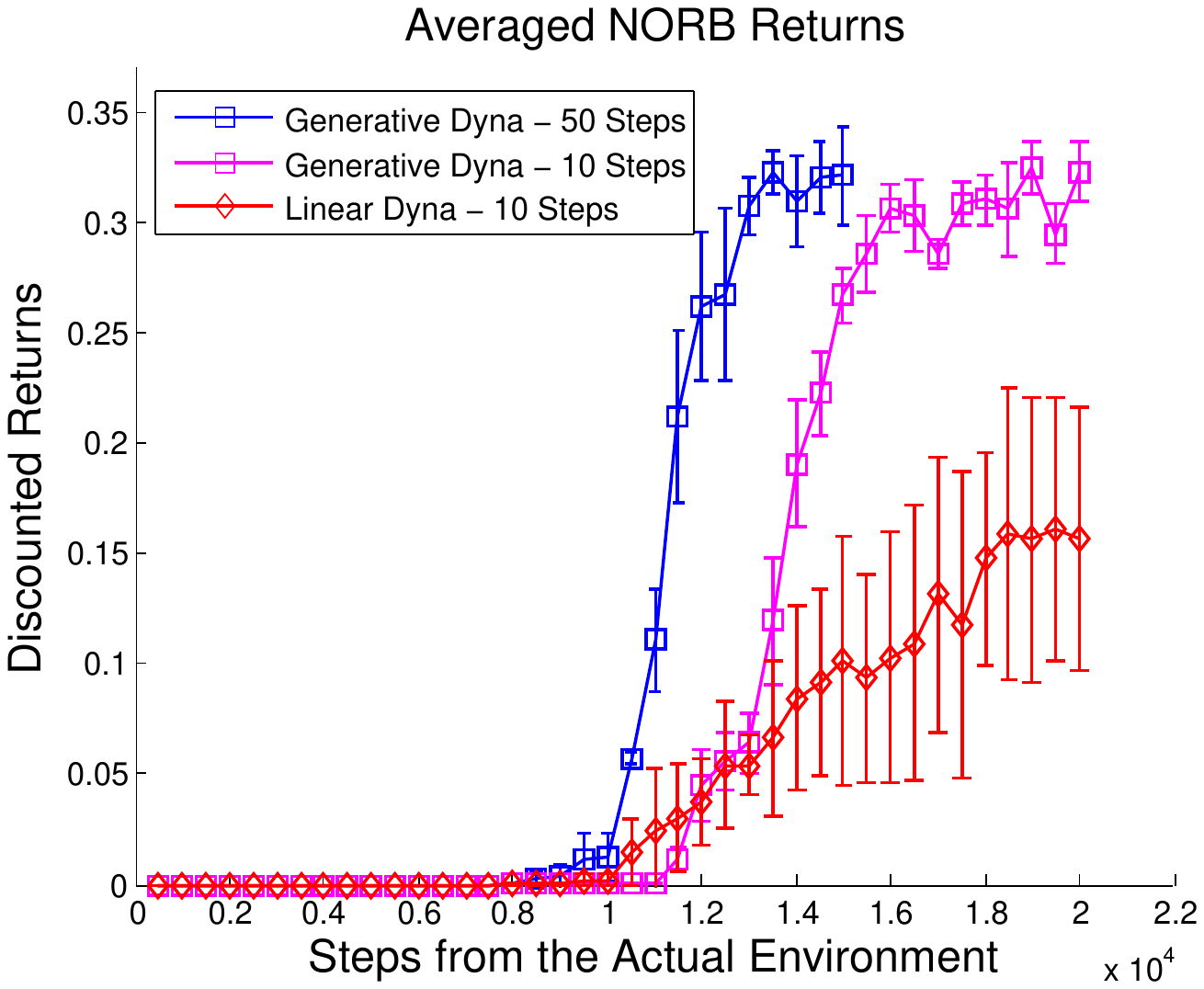} 	
	}	
	\vspace{-130pt}
  \centerline{
	\includegraphics[height=7.0cm]{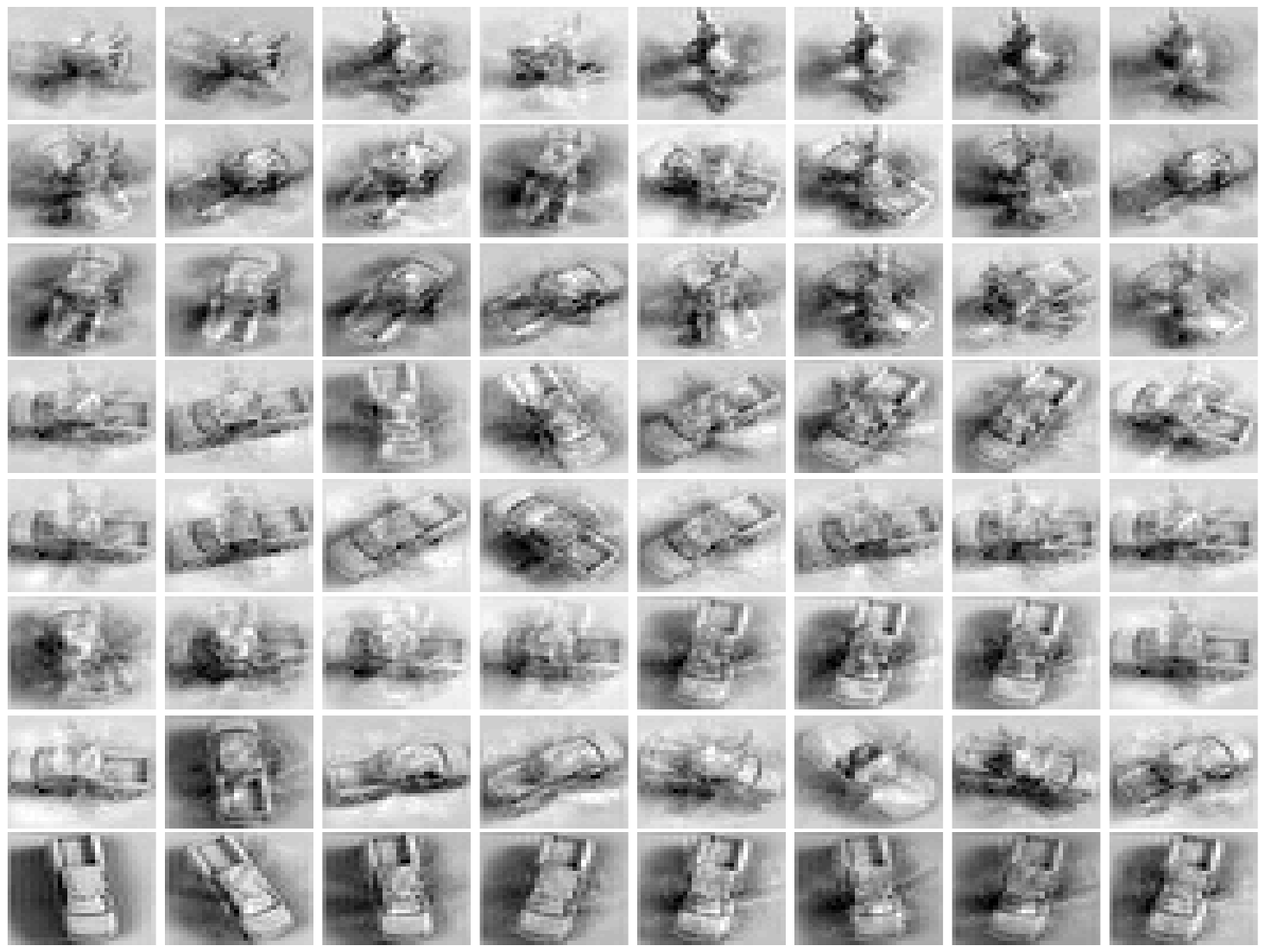} \hspace{-40pt}
	\includegraphics[height=7.0cm]{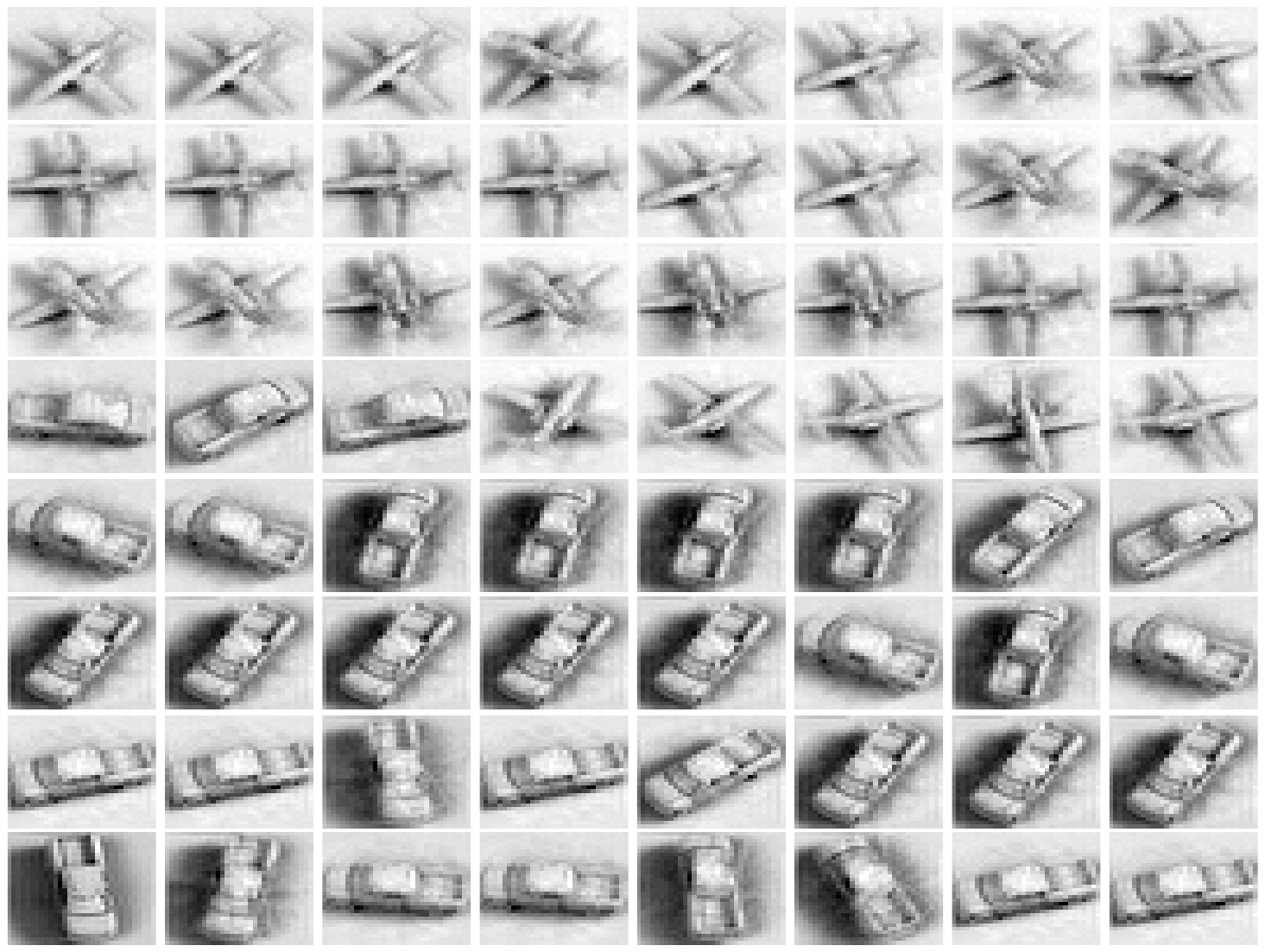} \hspace{-40pt}
	\includegraphics[height=7.5cm]{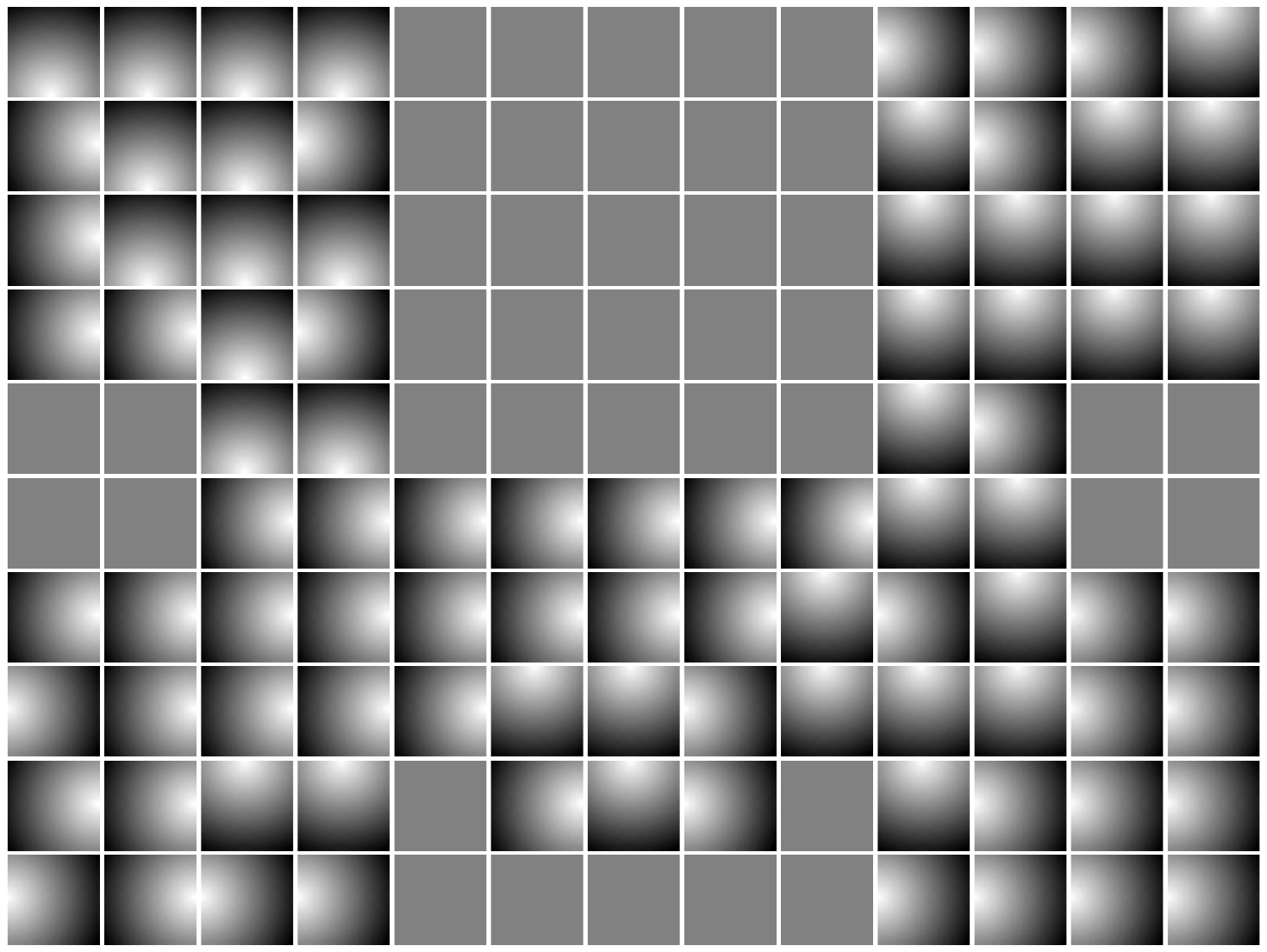}
	}
	\vspace{-60pt}
	\caption{\small{[Top] The averaged returns over the NORB environment. [Bottom] Sample walks over a linear (left) and generative (middle) Dyna models.  The walks begin in the bottom left cell and proceed left to right along the columns and up the rows.  At right is shown the policy learned using generative Dyna (bright spots indicate the direction taken in each grid square). }}
  \label{fig:norbResults}
\vspace{-10pt}
\end{figure}

The Dyna implementation was first compared with a baseline TD(0) / SARSA implementation over the environment - simply learning by sampling from the environment.  For every step taken in the actual environment $50$ simulated steps are used to further train the function appoximation model using the rule described in \S\ref{sec:background} with the function approximation parameters initialized to $0$.  The simulated states of the model are generated as detailed in \S\ref{sec:genPrediction}.  The Dyna samples contributed to learning with fewer episodes and less data overall (Fig. \ref{fig:norbResults}).  For the results depicted, Dyna returns were averaged over $10$ independent runs on five seperate models, trained independently.  The SARSA returns come from the average over $50$ independent runs.  Finally, the generative Dyna returns were shifted by $7200$ samples to account for the extra data used to train the models (however these samples may have been used for Dyna learning as well, but the generative model learns faster regardless).  Also shown (Fig. \ref{fig:norbResults}) is the policy inferred from one of the Dyna models.  Generative Dyna with $K=50$ at $7500$ steps over the environment took about $14$ hours total.

%
%

Next, we compared our results to those of Sutton et al. (2008) using a linear Dyna model.  This was accomplished by training a linear model using the documented implementation \cite{sutton2008dyna}.  The linear models were trained over the same data as the generative models with a learning rate of $0.001$ for $5$ sweeps over all of the samples, at which point the model weights had converged.  As before five independent models were trained with returns averaged over $10$ independent runs on each model and then averaged over each model.  Dyna was executed over the linear models using the same parameters as those used for Dyna with the generative model.  The plots in Fig. \ref{fig:norbResults} compare the returns for each model with different trajectory lengths.  It should be noted that the trajectory samples obtained from the linear models were of poor quality after a few steps (Fig. \ref{fig:norbResults}), this prevented us from executing trajectories longer than around $10$ steps, limiting the amount of accurate ``fantasizing'' that could be done with the linear model, and consequently its effectiveness as a simulator also.

\section{Conclusions and Future Work}\label{sec:conclusion}

Deep models provide a means in which to learn good high-level state representations for domains consisting of high-dimensional data.  Using these high-level feature representations further aids in learning a good temporal model of the environment along with timely prediction of future states.  The purpose of this paper is to demonstrate the benefits of implementing Dyna with non-linear generative models in comparison to the linear expectation models used in the state-of-the-art.  We showed this using DBNs however it is expected that there exist other non-linear models may be used to the same end; this is a suitable direction for future research.

The results of our experiments in \S\ref{sec:experiments} demonstrate two important properties of our approach: that the generative model of the environment can be trained with high accuracy, and that the model when used within the Dyna architecture can significantly speed up learning of the value function using linear function approximation.  We have also demonstrated that Dyna with generative models significantly outperforms Dyna implemented with linear models in a domain consisting of high-dimensional images with some variance in the state transition function.  Many problems in reinforcement learning derive from real world domains with partial observability and non-stationarity in the environment dynamics.  Thus, further investigation into how well this method handles changing time dependecies and more complex domains is a very important direction for future work.

%
\bibliographystyle{abbrv}
\bibliography{nipsBib}  
%
%
%
%
%

\end{document}